\documentclass[runningheads]{llncs}
\usepackage[T1]{fontenc}
\usepackage{graphicx}
\usepackage{orcidlink}
\usepackage{cleveref}
\usepackage{booktabs}
\usepackage[normalem]{ulem}
\usepackage{xcolor}
\newcommand{\etal}{\textit{et al.}}

\usepackage{dingbat}

\newcommand{\rot}[1]{\rotatebox{90}{#1}}

\begin{document}
\title{Entry Separation using a Mixed Visual and Textual Language Model: Application to 19th century French Trade Directories}
\titlerunning{Entry Separation using a Mixed Visual and Textual LM}
\author{B. Duménieu\inst{1}\orcidlink{0000-0002-2517-2058} \and
E. Carlinet\inst{2}\orcidlink{0000-0001-5737-5266} \and
N. Abadie\inst{3}\orcidlink{0000-0001-8741-2398} \and
J. Chazalon\inst{2}\orcidlink{0000-0002-3757-074X}}
\authorrunning{B. Duménieu et al.}
\institute{CRH-EHESS, Paris, France\\
\email{bertrand.dumenieu@ehess.fr}\and
EPITA Research \& Development Laboratory (LRDE), Le Kremlin-Bicêtre, France\\
\email{\{edwin.carlinet,joseph.chazalon\}@epita.fr}\and
LASTIG, Univ. Gustave Eiffel, IGN-ENSG, F-94160 Saint-Mandé, France\\
\email{nathalie-f.abadie@ign.fr}}
\maketitle
\begin{abstract}When extracting structured data from repetitively organized documents, such as dictionaries, directories, or even newspapers,
a key challenge is to correctly segment what constitutes the basic text regions for the target database.
Traditionally, such a problem was tackled as part of the layout analysis
and was mostly based on visual clues for dividing (top-down) approaches.
Some agglomerating (bottom-up) approaches started to consider textual information to link similar contents,
but they required a proper over-segmentation of fine-grained units.
In this work, we propose a new pragmatic approach
whose efficiency is demonstrated on 19\textsuperscript{th} century French Trade Directories.
We propose to consider two sub-problems:
coarse layout detection (text columns and reading order),
which is assumed to be effective and not detailed here,
and a fine-grained entry separation stage for which we propose to adapt a state-of-the-art Named Entity Recognition (NER) approach.
By injecting special visual tokens, coding, for instance, indentation or breaks,
into the token stream of the language model used for NER purpose,
we can leverage both textual and visual knowledge simultaneously.
Code, data, results and models are available at \url{https://github.com/soduco/paper-entryseg-icdar23-code},
\url{https://huggingface.co/HueyNemud/} (\texttt{icdar23-entrydetector*} variants).
\end{abstract}

\section{Introduction}

Discriminating semantically consistent content blocks,
such as entries in dictionaries or directories, or articles in newspapers,
is a challenging task for historical document analysis.
Most of the approaches proposed by the Document Analysis and Recognition (DAR) community
tackled the problem by extracting and analyzing visual clues,
mainly considering the problem as a refined version or, rather, the extension of the layout analysis problem.
This makes much sense, as proper block segmentation is a prerequisite,
but misses important textual clues.

In this work, we investigate the integration of visual features into an encoder-only transformer architecture (BERT here), originally targeted to text,
to form a multi-modal \emph{entry separation} system.
Our core idea and contribution consists therefore in injecting special tokens like, 
for instance, ``\texttt{<small\_left\_spacing>}'', ``\texttt{<large\_right\_spacing>}'', ``\texttt{<line\_break>}'', or ``\texttt{<column\_break>}'',
in the stream of regular tokens computed from the textual content
and process the resulting enriched stream with classical Natural Language Processing (NLP) techniques to predict split positions.

Such textual content is assumed to be extracted from documents like the ones illustrated in \cref{fig:directories},
i.e. a pretty clean document with a regular structure that a dedicated (visual only) preliminary layout analysis may segment correctly in columns.
This is an important, pragmatic assumption, as this intermediate representation level is, 
in practice, very stable in our experiments,
and, furthermore, enables us to create a multicolumn, multipage stream of text to later \emph{separate} into entries.
Please note that here we use the term \emph{separate} instead of \emph{segment}
to avoid confusion between the ``cutting'' of the text stream and the geometric decomposition of the page image.

This paper features the following contributions:
\begin{enumerate}
    \item We introduce a pragmatic approach to \emph{separate} semantically consistent parts of text (validated on directory entries)
    by combining visual and textual clues in a stream of text tokens, leveraging existing NLP techniques.
    \item We release a new public dataset under an open license to reproduce our results 
    and enabling the comparison of \emph{content separation} systems.
\end{enumerate}

After a survey of the techniques that can be used for a \emph{separation} task in \Cref{sec:relwork}, 
we detail our approach in \Cref{sec:approach}. We then introduce the new public data set and the metrics that we use to benchmark our \emph{separation} system in \Cref{sec:eval}. Lastly, we describe our experimental protocol and the results obtained in \Cref{sec:experiments}. \section{Related Works}
\label{sec:relwork}

Document Layout Analysis (DLA) approaches aim at identifying and labeling homogeneous regions of documents. DLA can be viewed as the first step in a document understanding pipeline. Subsequent stages, such as OCR or Named Entity Recognition (NER), process the document regions produced by this stage, and the quality of their results depends directly on the quality of the document segmentation done by the DLA step.  

\subsection{Top-down DLA strategies}
The survey of DLA approaches proposed by \cite{binmakhashen2019document} identifies three main strategies. Among them, top-down strategies consider the whole document level and split it into smaller regions based on some rules. They stop when there is no smaller region left or when some predefined stopping criterion has been reached.
Classical approaches are generally based on a variant of XY cuts~\cite{nagy1984hierarchical}.
This recursive cutting approach was improved with better preprocessing~\cite{ha1995recursive} and with a recovery of the reading order~\cite{meunier2005optimized,sutheebanjard2010modified}.
These approaches are very fast and can perform reasonably well on clean, well-separated document layouts.
However, they consider only superficial visual clues and cannot perform a fine-grained decomposition of a document
or classify regions precisely, as the semantics involved are unavailable to them.
Indeed, once some over-segmentation is reached, it is often necessary to reassemble document fragments using a dedicated process.
This is the case, for example, of the article separation technique for historical newspapers used in the NewsEye project~\cite{newseye2021}
which links blocks according to their content similarity to recover articles and reading order.

\subsection{Bottom-up DLA strategies}
Bottom-up strategies are the second type of DLA approach listed in \cite{binmakhashen2019document}. They start from small document features, such as pixels, words, or connected components, and group them based on some rules to create homogeneous and consistent regions until some predefined stopping criterion is reached. 
Hybrid strategies, which take advantage of top-down and bottom-up approaches, are the last type of strategies mentioned by \cite{binmakhashen2019document}.
The docstrum method~\cite{o1993document} is one of the first bottom-up approaches that has been proposed. It relies on a nearest-neighbor clustering of document components, based on measurements of skew, within-line, and between-line spaces, to identify the distinct regions. More generally, machine learning-based approaches that classify document elements according to regions can be viewed as bottom-up approaches. Many traditional machine learning approaches have been leveraged, either directly on pixel intensities \cite{chen2014page,fischer2014combined}, on hand-crafted features based on Gradient Shape Feature (GSF) \cite{diem2011text} or Scale Invariant Feature Transform (SIFT) \cite{garz2011layout,garz2010detecting}. These visual feature-based solutions have yielded better results than the former. More recently, deep learning-based approaches have become popular \cite{chen2017convolutional,gruning2019two,oliveira2018dhsegment}, establishing new baselines for many DLA applications. Contrary to classical machine learning methods, they generate their own visual features based on pixel data to classify pixel into document regions.
Hence, such methods can form homogeneous regions with unconstrained shapes, and sometimes classify them at the same time.
However, the classes remain rather limited, and these methods are preferably used as a preprocessing to form stable regions which will be later processed according to specific rules. 

\subsection{Multimodal DLA approaches}
Early deep learning-based DLA approaches have focused on visual information only to solve the most common DLA tasks. \cite{yang2017learning} innovated by proposing a fully multi-modal convolutional network, involving also a text embedding map to perform a page semantic segmentation task. \cite{liu-etal-2019-graph} handles texts segments as graph nodes and uses a GCN to compute text segment embeddings capturing their visual and textual context. These embeddings are then used jointly with text embeddings in a key information extraction task.   
The recent advent of Transformer-based models has led to the development of multimodal approaches, which jointly leverage visual, textual, and layout information in the self-supervised pretraining step. The LayoutML~\cite{xu2020layoutlm} approach also includes, in addition to a textual embedding, two additional types of embeddings for the input tokens in the original BERT~\cite{devlin2018bert} architecture: 1) a 2D position embedding representing each token bounding box coordinates in the document local coordinates system, provided by the OCR preprocessing step, and 2) a visual embedding represented by an image patch of each token. The former is thus intended to capture the tokens relative positions in the document, and the latter its visual properties (color, font, etc.). The model is then pretrained on two self-supervised tasks. For the Masked Visual-Language Model task, the model has to predict masked tokens based on their context tokens and 2D position embeddings. In the Multilabel Document Classification task, the model has to predict each document category using all the available embeddings. The model is finally fine-tuned for specific DLA tasks (form understanding, receipt understanding, and document classification) using well-known annotated benchmark datasets. The LayoutMLv2~\cite{xu-etal-2021-layoutlmv2}, LayoutMLv3~\cite{huang2022layoutlmv3}, StructuralML~\cite{li-etal-2021-structurallm}, XYLayoutLM~\cite{gu2022xylayoutlm}, BROS~\cite{hong2022bros}, and LayoutXLM~\cite{xu2021layoutxlm}, LiLT~\cite{wang-etal-2022-lilt} approaches build on this first proposal and propose new embeddings, attention mechanisms or pretraining tasks to improve the model's ability to take into account text, layout and visual content at the same time.
In practice, these approaches produce state-of-the-art results for region tagging, but at the cost of a training with many annotations, including object positions.
Furthermore, their architecture is not well suited to capture complex textual structures,
and it is necessary to use a dedicated network as a downstream step to extract structured information.
The Document Attention Network~\cite{Coquenet2023} is an end-to-end DLA, OCR, and structured information extraction system that produces a structured output directly from a document image.
Such an approach is very promising but still suffers from some limitations.
It is complex to train and can not yet deal with a series of pages at the same time.
Furthermore, it is not possible yet to provide corrections at an intermediate step, like fixing a column position,
and may not offer the modularity we expect from a system dedicated to the processing of historical documents,
where both the capability to reuse existing, off-the-shelf component and the ability to be corrected (to propose an assisted digitization process) are of high importance.
As a result, dividing the content separation problem into two sub-problems seems a pragmatic, but unexplored approach for mostly textual documents with a rather clean and simple structure:
1. use an existing DLA stage for a coarse segmentation and extract a textual stream;
and 2. use a dedicated text processing stage to isolate and tag relevant content.

\subsection{Named Entity Recognition in Text Sequences}
Named Entity Recognition (NER) \cite{nadeau2007} is a common natural language processing task that aims to locate and classify portions of text that represent entities of predefined categories such as people, places, organizations, dates, etc. The early approaches were mainly based on hand-crafted rules designed to retrieve lexico-syntactic patterns \cite{nouvel2011} or dictionary entries \cite{maurel2011,mansouri2008} representing named entities. With the use of a comprehensive dictionary and significant efforts in rule engineering, these approaches give good results for clean, domain-specific texts~\cite{nadeau2007}. 
Supervised approaches include traditional machine learning approaches based on carefully chosen text features and deep learning approaches that automatically build their own features for classification from raw text. As highlighted in the survey proposed by \cite{li2020}, in recent years, deep learning-based NER approaches have outperformed both supervised and unsupervised feature-based approaches, achieving state-of-the-art results. BERT-based approaches (for Bidirectional Encoder Representations from Transformers)~\cite{devlin2018bert}, use the Transformer attention model \cite{vaswani2017attention} for language modeling. They take advantage of the transfer learning principle: training a neural network on some known generic tasks ---~here Masked-Language Modeling and Next Sentence Prediction~--- and then fine-tuning it for a more specific purpose. To fine-tune a BERT model for the NER task, one only needs to feed the output vector of each token into a classification layer that associates a label to it. Currently, transformer-based models give the best results for the NER task \cite{li2020}.
Recently, Hao et al.~\cite{hao_language_2022} showed that language models were general interfaces that can merge multiple modalities.
Modern transformer libraries make it easy to add and use custom tokens.
While using images embedding directly might be a possibility, directly generating tokens based on visual properties of the text sequence has not yet been proposed, to our knowledge.

\subsection{DLA tasks and benchmark datasets}

Many datasets have been proposed by the community to evaluate DLA approaches. They are designed to train and evaluate models on one, two, or three well-targeted task types. For example, the FUNSD~\cite{jaume2019funsd} and XFUN~\cite{xu2021layoutxlm} datasets are designed for form understanding, which includes text field detection, OCR, semantic annotation, and linking of document regions. RVL-CDIP~\cite{harley2015evaluation} is made up of images of documents of various types, annotated to classify them into categories. SROIE~\cite{huang2019icdar2019} and CORD~\cite{park2019cord} provide scanned images of annotated receipts for text localization, OCR, and key extraction tasks. DocVQA~\cite{mathew2021docvqa} combines document images, questions, and their expected answers for a VQA use case. SciTSR~\cite{chi2019complicated} contains several thousand tables extracted from scientific articles and annotated for a structure extraction task. Some of these datasets, such as XFUN, have documents written in various languages.

The set of scanned and annotated newspaper articles proposed by the NewsEye project for the ICPR 2020 competition on text block segmentation~\cite{newseye2021} shows to correspond the most to our goal of separating directory entries. More specifically, the dataset provided for the ``simple'' task, which only deals with regular multicolumn newspaper pages, without images or tables, has a layout that looks like directory or dictionary entries. Moreover, the task to be performed, which consists in grouping the detected lines of text into semantically consistent paragraphs, has a purpose similar to that of separating directory entries. But directory or dictionary entries are short texts written according to a well-defined pattern, where case, punctuation, line breaks, and indents play a strong structuring role. Newspaper pages, even without images, include more visual information to organize the articles, such as blank spaces or horizontal and vertical lines. Therefore, it seems useful to produce and publish an open benchmark dataset for entry separation in this type of document.

 \section{A Mixed Visual and Textual Language Model}
\label{sec:approach}
We detail in this section our proposal for a mixed \emph{visual and textual} language model.
We consider it as a sort of response to the CharGrid~\cite{katti_chargrid_2018} approach from the point of view of natural language processing (NLP).
CharGrid introduced an elegant way to process administrative documents, such as invoices, using both textual and visual information.
It did so by projecting textual information (simplified character codes) into the image domain:
pixels were ``colored'' according to the bounding box of the character they fall into.
This enabled the authors to use classical Computer Vision (CV) object detection techniques based on the Faster-RCNN architecture~\cite{ren_faster_2015}.

This was an interesting case of the Document Analysis and Recognition (DAR) community benefiting from CV advances.
In the case of single-page documents whose information is mostly carried by the spatial organization of contents, this approach produced great results.
However, for documents whose information is carried mainly by textual content, such as the trade directories we introduced earlier (\cref{fig:directories}), such an approach is limited, and it is necessary to consider approaches that emerged from the field of NLP such as BERT~\cite{devlin2018bert} to capture rich syntactic structures.
Like Katti \etal{} showed how to leverage Faster-RCNN for DAR, we propose a way to leverage BERT (an encoder-only transformer which implements a non-causal language model) for DAR.

We propose a method where visual tokens are injected in the regular token stream (usually built from text content) and capture extra visual clues which can assist the network in discriminating and tagging input's fragments as expected.
\\[0.5em]

Our proposed approach is composed of the following steps:
\begin{enumerate}
    \item \emph{Run a \emph{coarse} document layout segmentation.}
    In this paper, we assume that the performance of this preliminary step is strong enough to further process its results.
    In practice, we found that for mostly textual documents whose structure is relatively simple and clean, some ad hoc, top-down image splitting followed by OCR's line detection performs well.
    We use the line extraction algorithm from the same OCR system that we use in step 2 because OCR models are trained on data produced by some specific line extraction algorithm, and we witnessed performance degradation when using other detection techniques.
    A single example line may look like this, with its enclosing polygon drawn in dashed red:\\
    \fbox{\includegraphics[height=1.5em]{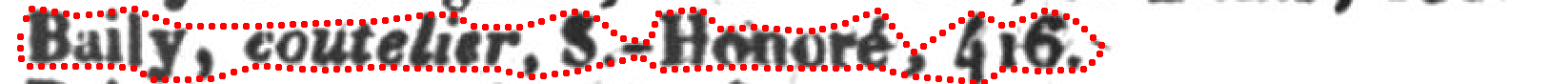}}

    \item \emph{Run character recognition on the set of pages we intend to process.}
    We store the transcription for each detected text line, along with its exact position on the page and the blocks it belongs to.
    In the experiments we report here, we only kept column information and ran the OCR system column-by-column.
    In our previous example, the following (noisy) text may be obtained:\\
    \fbox{\texttt{"Bailly, coutelier. S.Honuré, 416."}}

    \item \emph{Compute extra visual clues which will be merged into the text stream as extra tokens.}
    Visual clues can be of several kinds and will be detailed hereafter.
    In the experiments we report here, we add new tokens to represent breaks for line or columns, left-hand line indentation, and right-hand horizontal space left at the end of a line.
    In our running example, the following items would be generated, omitting their position attributes for clarity:\\
    \fbox{\texttt{[<LSPACE size=0.0056/>, <RSPACE size=0.3490/>, <LINE\_BREAK/>]}}

    \item \emph{Prepare a categorical encoding of parametrized tokens.}
    Some tokens may carry numerical values, such as the length of horizontal spaces, and we need to perform a categorical encoding before using them as inputs.
    We know that some techniques directly use numerical tokens, but we did not experiment them.
    In our running example, the categorical tokens may look like:\\
    \fbox{\texttt{[<LSPACE\_SMALL>, <RSPACE\_LARGE>, <LINE\_BREAK/>]}}

    \item \emph{Merge visual and textual content in textual representation.}
    In practice, merging visual and textual clues boils down to adding a few special text strings at the right position in the input text stream, and configuring the tokenizer with the new tokens and their associated special strings.
    In our running example, the resulting text content would look like this:\\
    \fbox{\texttt{"<LSPACE\_SMALL>Bailly, coutelier \dots{} <LINE\_BREAK/>"}}

    \item \emph{Tokenize the textual representation.}
    Text tokenization is a necessary step to capture frequent text patterns and build better NLP models.
    Input text undergoes some chunking process, where each chunk is assigned the identifier of the pattern it matches with. Modern NLP libraries make it easy to add new tokens from special substrings.
    First, we can define rules to substitute some arbitrary text from the original text stream with a special token,
    and second we can easily change the weights of the input layer of the BERT transformer to deal with new tokens,
    without changing already-trained weights assigned to other tokens.
    In our running example, tokenization can look like this, with the numbers between brackets being token unique identifiers:\\
    \fbox{\texttt{[<LSPACE\_SMALL>(1001), "\_Bail"(118), "ly"(386), ","(10) \dots{}]}}

    \item \emph{Process the token sequence using some BERT-like, encoder-only transformer.}
    We implement a token classification task which assigns a label to each token.
    Possible labels and how they can be assigned to tokens is detailed hereafter.
    This last step assigns categories to text spans or isolated tokens.
    This enables us to split and tag the input stream simultaneously.
    Another advantage is that it is possible to reuse a pre-trained network and fine-tune it for a particular document type with a limited annotation cost:
    annotating text splits is very fast and easy, as it only consists in inserting tags in text.
    Also, the amount of annotation required to train a network is within reach of a small team (as the experiments show), and the computation cost of a text transformer, both in terms of training and inference, is well suited for digital humanities applications.
    In our running example, some possible labeling for simultaneous entry separation and NER can be:\\
    \fbox{\texttt{[EBEGIN, I-PER, I-PER, O \dots{}]}}
\end{enumerate}

\paragraph{Inputs tokens.}
The approach we propose enables the input stream to contain a sequence of ``physical'' tokens of several kinds.
(We indicate in \textbf{bold} the tokens we use in the experiments we report in \cref{sec:experiments}.)
\begin{itemize}
    \item \textbf{Regular \emph{text tokens}} produced by the tokenizer of the NLP model: they belong to a previously trained (context-free) lexicon, according to the SentencePiece~\cite{kudo_sentencepiece_2018} tokenization approach. The CamemBERT model that we used was trained with these particular tokens.
    \item \emph{Structural tokens}: \textbf{left/right space}, \textbf{line, column and page breaks}, vertical spaces, etc.
    \item \emph{Graphical tokens}: horizontal rules, decorations, bullets, etc.
    \item \emph{Presentational tokens} in the form of pairs of start and end tags surrounding the content described: text size, text style, text family, etc.
\end{itemize}
Such an approach is very flexible.
The only delicate part is to identify the correct position where each token should be inserted in the text stream.
For documents whose content is mostly textual, this information is easily derived from the coarse structure, the reading order, and the polygons of the text lines.
It may also be suitable to process digital native documents like PDFs, where reading order is recovered using coarse layout analysis, and style attributes are easily obtained.

\paragraph{Output labels and annotation policies.}
The output stream contains the sequence of predictions, assigning a label to each element of the input stream.
We use matching pairs of labels to indicate the beginning and the end of the semantic constructs we intend to detect.
Even if assigning a ``cut'' label to a particular token is possible, for instance, to mark separation points in a stream of directory entries,
we believe that a pair-of-tags approach is safer in production.
Indeed, real input streams may contain unexpected structures, and we want to be able to discard them from our automated extraction,
rather than just partitioning the input.
It is also possible to perform content separation (of headings, entries, page numbers\dots) and structured content extraction from entries at the same time, using the same network.
This is a very attractive feature both in terms of computation efficiency (the same network is called once to produce several outputs)
and ability to leverage more training annotation: extents of headings and entries at one level, and of key entities in entries at another level.
However, deciding which token will carry each label is not straightforward, and several strategies are possible.
Using joint labeling or any hierarchical classification technique, more than one label may be assigned to a given input token,
but this technique requires some special label preparation or a custom architecture.
Another option is to take advantage of the extra ``visual'' tokens and label them with some structural tags.
For instance, the tokens for left- and right-line spaces may be used to annotate beginning and end of entries.
Note that when we use such tokens, they are always present even if the space is null: in the particular case of the left space indent, we would use several tokens in the categorical encoding of the possible space sizes, reserving one token for zero-width space.

 \section{Evaluation Framework}
\label{sec:eval}
We present here the dataset and metrics used to assess the performance on real, noisy OCR text of the entry separation task and the NER task.
The NER task may be run in conjunction with the entry separation task.

\subsection{Dataset}
\label{sec:eval-dataset}

\begin{figure}[tb]
    \center{\includegraphics[width=0.9\textwidth]
        {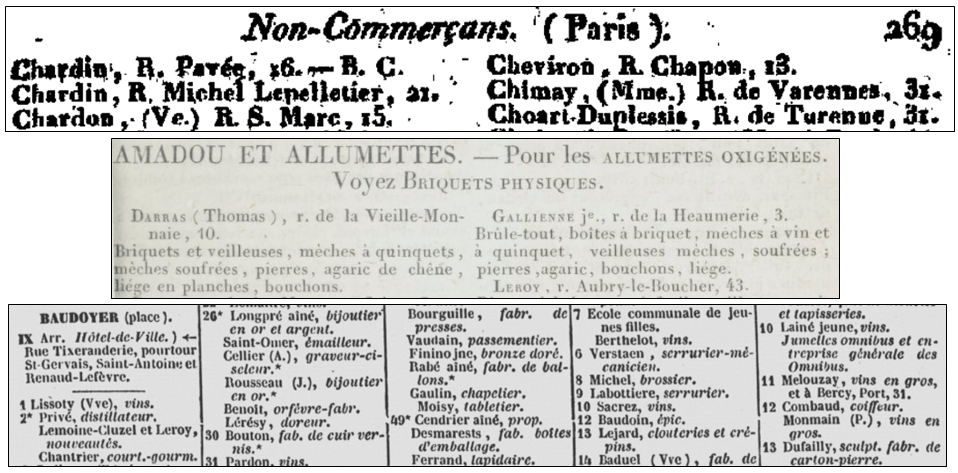}}
   \caption{\label{fig:directories} Examples of directory layouts and contents: 1) Duverneuil et La Tynna 1806 - index by name; 2) Deflandre 1828 - index by activity ; 3) Bottin 1851 - index by street name}
\end{figure}

In this work, we assume that a general-purpose segmentation technique can be used to detect text columns reliably; hence, we
will not measure the performance on this stage on the pipeline and instead consider that text columns are already
properly segmented. Of course, such assumptions can cause a cascade of errors when the first stages of a pipeline fail,
but in our experience, such segmentation is quite reliable on printed contents like the ones used hereafter.
The dataset used in the experiments is augmented from the 19\textsuperscript{th} Paris trade directories dataset
from~\cite{abadie.das22}. The directories are available from different libraries in Paris and have been independently digitized at various levels of quality. Directories contain social information and localization of people that
are mainly organized into lists.
For instance, the entry ``\textit{Batton (D.-A.) \includegraphics[height=1em]{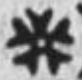}, professeur au
Conservatoire de musique et de déclamation, Saint-Georges, 47.}'' from the \emph{Didot (1854)} contains the person's
name, his surname, his title (\emph{Légion d'Honneur}), his activity (professor at the Conservatory of music and
declamation), the street number (47) and the street name (Saint-Georges) of his living. This information is part of the
annotations provided by the dataset.
The dataset covers a long period of time, and the directories originate from several publishers. Therefore, their contents
(layout, methods of printing) vary greatly and the formatting of entries as well (see fig. \ref{fig:directories}). Even
within the same directory, the indentation rules may vary from page to page.
Therefore, our experiments focus on the task of properly placing directory entries \emph{separators} in the token stream we
generate. As previously explained, this stream can be composed of tokens of different nature, depending on whether we
add special visual tokens and on how we decide to model such visual information.

The dataset consists of 80 annotated pages, containing a total of 8\,900 directory entries.
Entries are made of 1.4 text lines on average, and most entries are single lines.
This means that a dummy classifier which predicts entry separators for each line would reach an F-score of 83.3\% (metrics are detailed hereafter).
80\% of the pages are assigned to the train set, 15\% for the test set, and the remaining 5\%
are used for validation.

Results we report are computed from noisy OCR inputs, even if the reading order is assumed to be correct.
This means that we projected the target entry separation and entity tags over a steam which contains noisy OCR content,
while such annotation was made on clean text.
Several tools are available to project NER targets, or generally tag positions, annotated on manually corrected OCR text onto noisy, raw OCR results~\cite{abadie.das22,nerval}.
When OCR results are very poor, annotation alignment tends to fail and related items are discarded.
Hence, this evaluation protocol reports results for acceptable OCR noise, and only measures errors introduced by the stream annotation stage we propose.
Errors due to upstream stages should also be measured to estimate the global performance of such a system.

\subsection{Metrics}
We consider the entry separation problem, as well as the NER problem, as token classification problems.
This makes the evaluation very simple as we consider a prediction to be correct only if it exactly matches the position of some expected one.
For the entry separation task, only one token is labeled for each ``entry begin'' and ``entry end'' markers.
For the NER task, we predict start and end positions similarly for each entity, and positions must match exactly to be considered as correct.
This evaluation protocol has the advantage of not making any assumption on the importance of small errors:
all errors are treated equal, and in the particular case of directory entries, this makes much sense, as a lost separator is a lost entry.
Therefore, we report standard classification metrics such as precision, recall, and F-score.
For entry detection, we consider two classes: entry-start cuts and entry-end cuts,
and metrics are computed in a straightforward way, as only one token can carry the correct information.
For NER, each entity is considered as a sample and metrics are reported accordingly, disregarding the number of tokens it contains.
We use the SeqEval~\cite{seqeval} library to compute our metrics.
\\We are aware of the evaluation process used in the ICPR 2020 Competition on Text Block Segmentation on a NewsEye Dataset~\cite{michael_icpr_2021},
which internally uses the coverage metric of the READ-BAD evaluation framework~\cite{gruning_read-bad_2018}.
We believe that this metric, which was designed to assess the quality of text baseline detection,
is too permissive, as it tolerates small boundary variations and fragmentation.
It also requires a matching step, which seems superfluous in our case.
 \section{Experimental Validation}
\label{sec:experiments}

\begin{figure}[tb]
    \includegraphics[width=\linewidth]{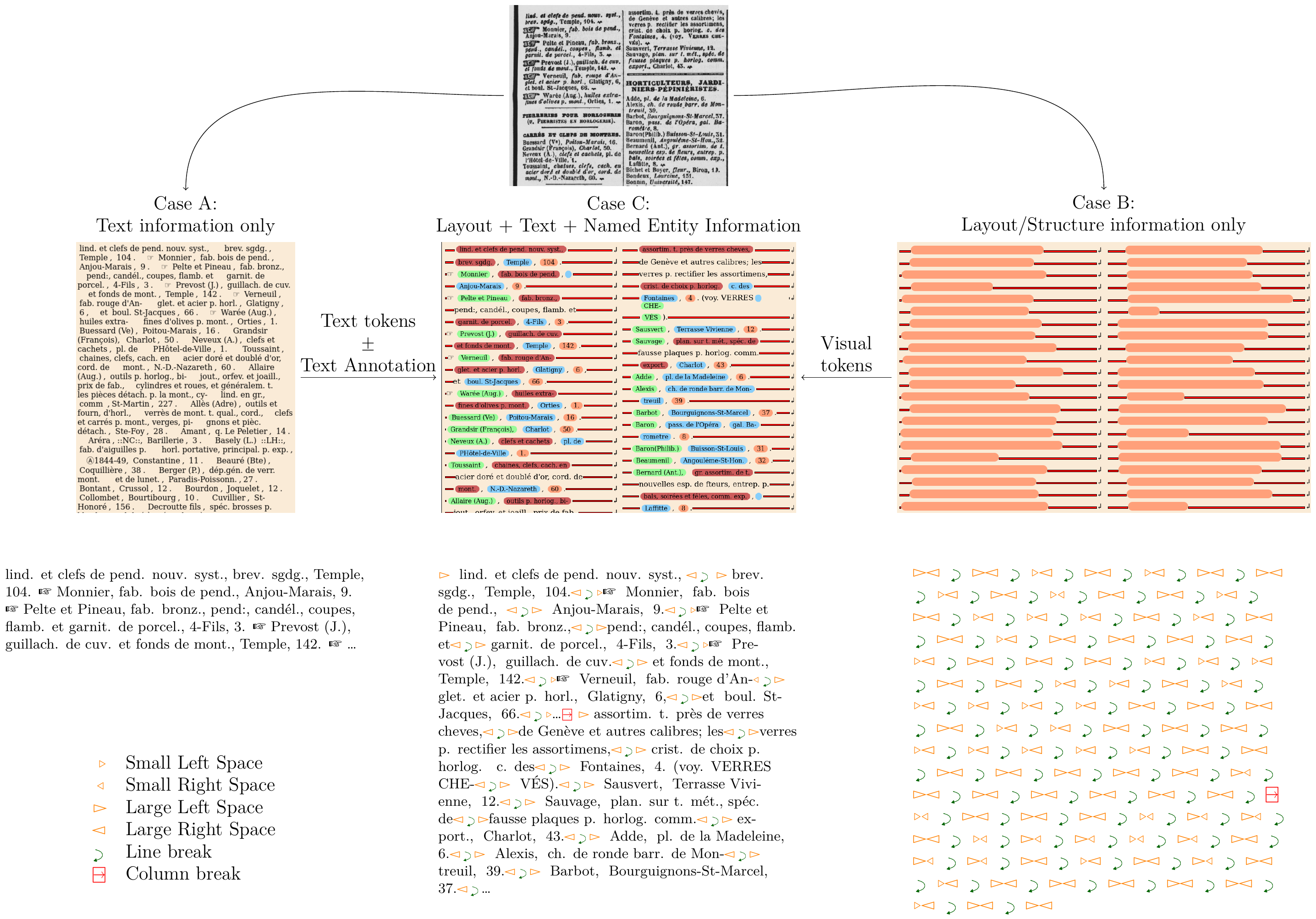}

    \caption{Illustration of possible input data.
    In all cases, input data is stream of tokens from which we have to detect \emph{entry} delimiters.
    In the left case (A), only the textual information (no line breaks, column breaks\ldots) is used, as a pure text token stream.
    On the other side (case B), only the visual layout information is used with \emph{breaks} and left/right blank spaces of the lines encoded as categorical tokens.
    The middle case (C) shows the combination of both kind of information, plus the annotations of the text entities: names, cardinal, location\ldots
    Below each case, we show the stream of tokens that is passed to the system for detecting entries.
    }
\end{figure}

In this section, we describe the experimental setup we use to demonstrate the feasibility and performance of our proposed approach.
As already mentioned, we focus on processing noisy text extracted by OCR, once coarse layout is detected.
In the case of trade directories, this text content represents directory entries in the order they appear, across rows, columns, and pages. Among this continuous stream of text, we first want to identify tokens that delineate directory entries.
Then, in a second step, we seek to jointly identify these separation tokens, as well as those that compose named entities.
We thus have two tasks to perform: one to recognize tokens that delineate text structure (entry separation task), and the other to recognize semantic information (NER task).
In the rest of this section, we make explicit which extra tokens are used, detail the progress of the experiments we conducted, and discuss their outcomes.

\subsection{New tokens definitions}
To help the language model recognize the entry beginning and end, we inject new tokens into the tokenizer's vocabulary.
\texttt{<break>} which represents all breaks in the text stream like new lines, column breaks, and page breaks, 
and \texttt{<lhspace-x>} and \texttt{<rhspace-y>} which represent left- and right-line indents detected by the coarse layout detection algorithm.
$x$ and $y$ are integers that quantize the value of the indent.
We measure indent in two ways: 1.~its width expressed as a percentage of the total line width (referred to as \emph{`absolute'} values),
and 2.~the width difference with the previous indent on the same side of the text (\emph{`relative'} values).

\begin{figure}[tb]
    \centering
    \includegraphics[width=0.45\textwidth]{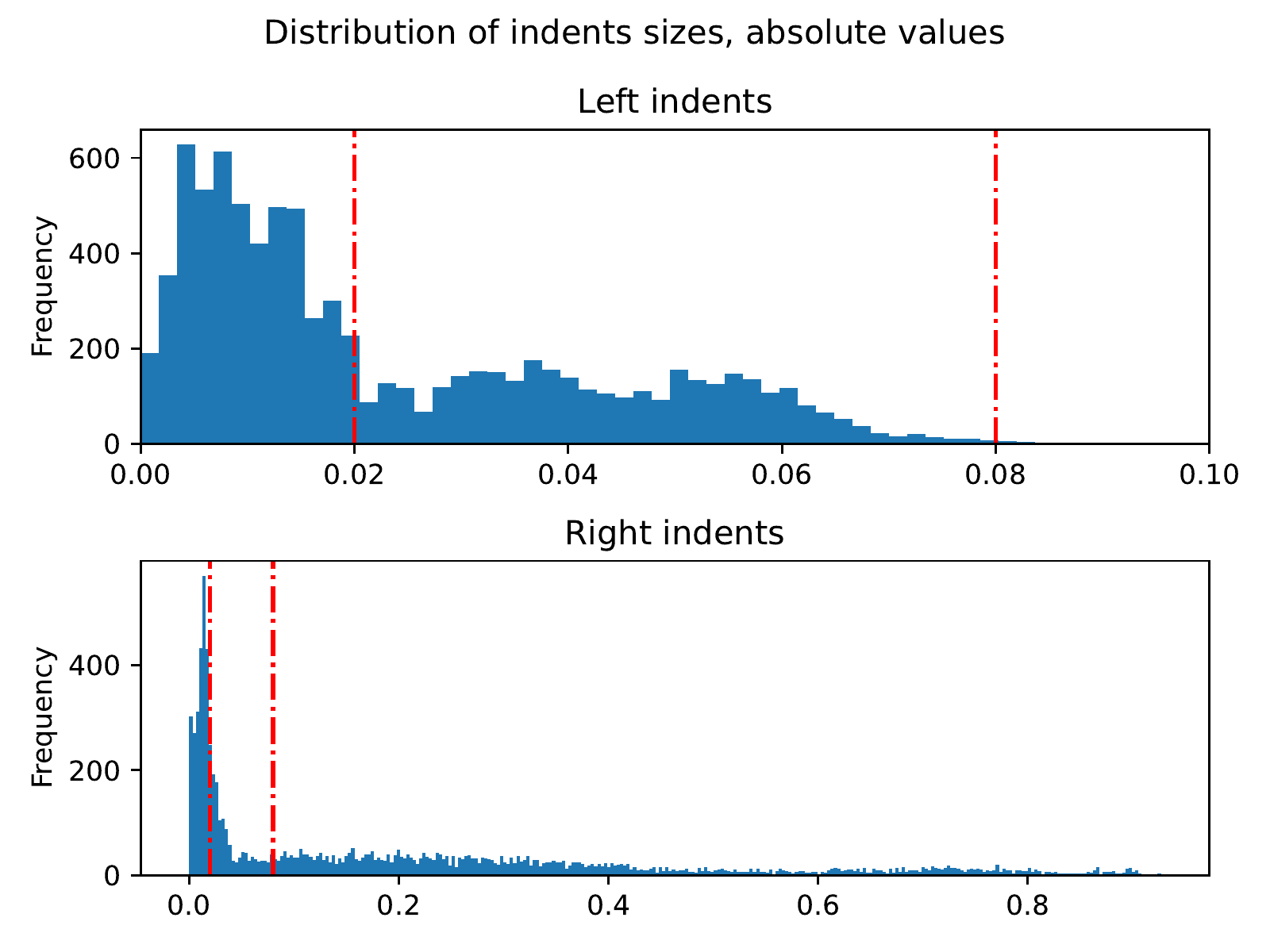}
    \hfill
    \includegraphics[width=0.45\textwidth]{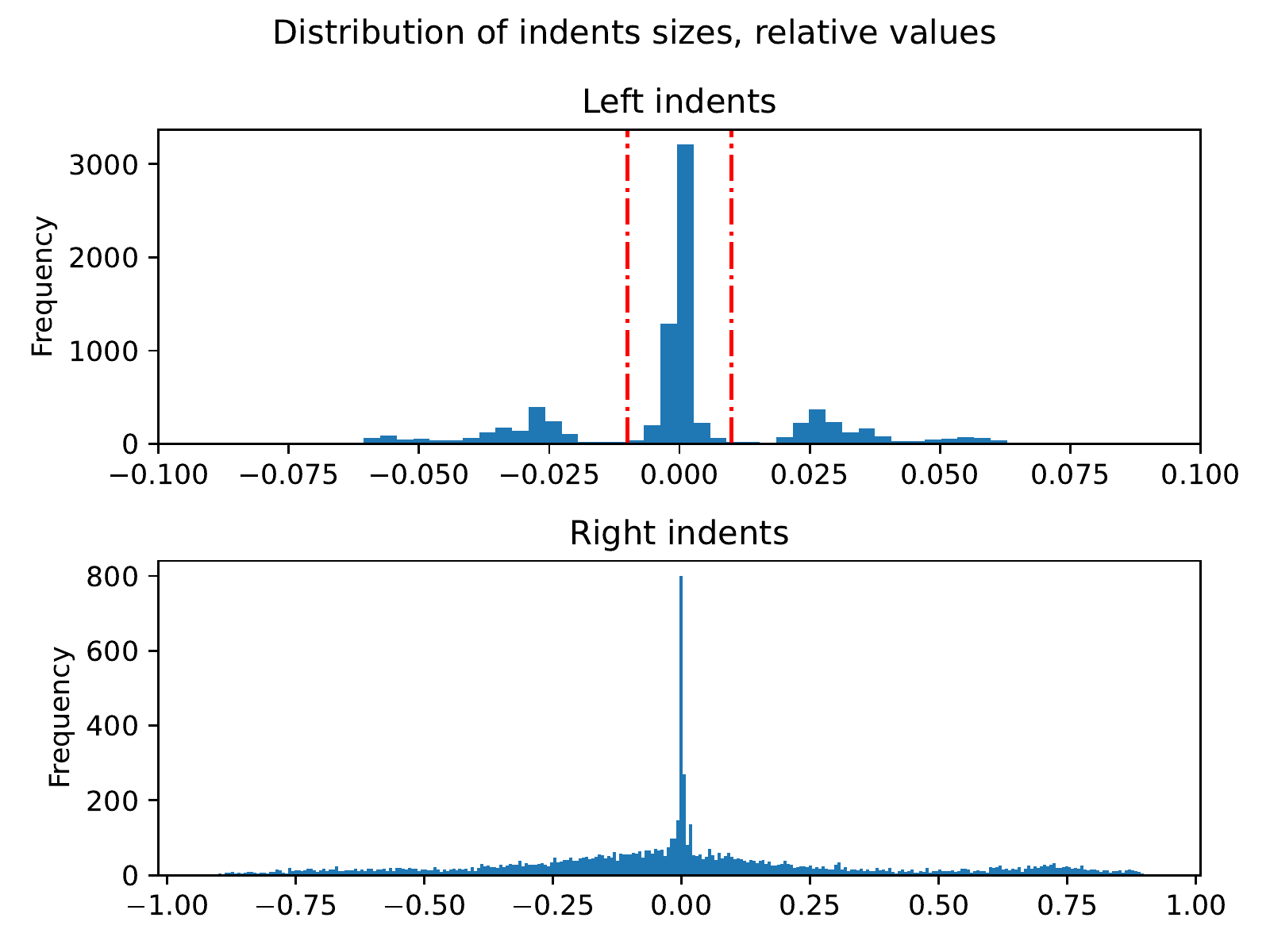}
    \caption{Distributions for left and right spaces on the training set.
    \emph{Left}: values for absolute measures.
    \emph{Right}: values for measures relative to the previous line.
    Right relative spaces were not considered in our experiments.
    Red dashed vertical bars indicate thresholds used for categorical encoding.
    }
    \label{fig:spaces_distrib}
\end{figure}

We performed a categorical encoding of left and right spaces, both for absolute and relative variants, before feeding them to the tokenizer.
The threshold values used were determined manually in the training set exclusively.
Distributions for absolute and relative spaces are displayed in \cref{fig:spaces_distrib}.
The values of the relative left indents are binned into $]-\infty;-0.01[$, $[-0.01;+0.01[$, and $[+0.01;+\infty[$ to produce three categorical tokens.
Similarly, the values of the absolute left indents are binned into $[0.00;0.02[$, $[0.02;0.08[$, and $[0.08;1.00]$,
and the values of the absolute right indents into $[0.00;0.05[$, $[0.05;0.08[$, and $[0.08;1.00]$.
We aimed at producing a small amount of tokens to ease model's training.
For absolute spaces, our logic was to produce three tokens: 
one for clear \emph{``no indentation''} 
another for clear \emph{``indentation''} cases, and the last for ambiguous cases.
For relative spaces, we also tried to capture 3 simple configurations:
\emph{``no significant change''}, \emph{``significant negative difference''} and \emph{``significant positive difference''}.

Finally, we also used a \texttt{<textline>} token which represents all the textual content found between two consecutive indents.
This particular case is intended to replace the actual textual content in the input to assess the performance of visual-only separation, with minimal alteration of the network.

\subsection{Experiments}

To assess the interest of each injected visual feature, we add them one by one and compare the results obtained by each
strategy. Then, we perform a second set of experiments to jointly recognize entry boundaries and named entities in the text. An
entry boundary is modeled with two dedicated NER labels: EBEGIN and EEND. In all the following experiments, the entry
boundary labels are marked on text tokens only, except for experiment 1.7 and 2.1.

The first set of experiments (xp-1.X) explores the ability of a token classification model to recognize entry boundaries with a varying
number of visual information tokens injected in the text and different configuration of indents. It starts with the raw
text stream (xp-1.1) and adds more and more visual clues. Experiment 1.2 adds visual break separators, at the line,
column or page level, and experiments 1.3 to 1.6 add space tokens to express the indentation level. We propose to
assign two kinds of values to those spaces: 1. the (\textbf{A}bsolute) distance between the line borders and the column
borders (as a percentage of the column width), 2. the (\textbf{R}elative) offset of the current line borders
with the border of the previous line (also as a percentage of the current line.). Although the first valuation represents
the level of indentation or if a line reaches the right border of the column, the \emph{relative} approach could eventually
detect a break in the alignment. In experiment 1.7, we study the opposite approach of xp-1.1, where we evaluate the
system only on the visual content (all spaces and breaks, but no text). The following table summarizes the information
embedded in the input for each experiment.

\def\yes{\checkmark}
\begin{center}
\begin{tabular}{llllllll}
    \toprule
    Experiment    & 1.1   &   1.2 &   1.3 &   1.4 &   1.5 &   1.6 &   1.7 \\
    \midrule
    Text          & \yes  &  \yes &  \yes &  \yes &  \yes &  \yes &       \\
    Visual breaks &       &  \yes & \yes  & \yes  & \yes  & \yes  & \yes  \\ 
    Left spaces   &       &      &   A   &  R    &  A    &  R     &   R    \\ 
    Right spaces  &       &       &       &       & A      & A      &   A   \\ 
    \bottomrule
\end{tabular}
\end{center}

In experiments 2.X, we propose to train the model for both named entity recognition \textbf{and} entry separation at the same time.
The two extra labels, EBEGIN (EntryBEGIN) and EEND (EntryEND), are added to the set of labels to
predict by the NER. Experiment 2.1 is thus very similar to xp-1.6, but those labels are set on the left and right
spaces. In contrast, in experiment 2.2, we propose to evaluate the system on a joint labeling of the first and
last words of the entry (which can, for example, be \texttt{(ACT, EBEGIN)}).
Such approach relies on new labels composed of the product of entry separation marks and NER labels.

\subsection{Training protocol and parameters}
All experiments were run 3 times with different random seeds.
We report the averaged results over these runs.
The base network is the public CamemBERT variant from~\cite{abadie.das22} pretrained on noisy OCR text from the same trade directories using an unsupervised masked language modeling task.
This network was not trained for NER on trade directories, but was able to capture some lexical and syntactic structures of the dataset during pretraining.

For each experiment, the network was trained with batches of randomly cropped sequences of tokens of variable sizes.
We generated these tokens according to each of the variants presented earlier in this section.
Each network was trained with up to 7500 training steps, and the F score was evaluated every 300 steps.
Training was stopped when this metric did not improve for 5 consecutive evaluations.
We use the AdamW optimizer as recommended, with a learning rate of $10^{-4}$ and a weight decay of $10^{-5}$.

\subsection{Results and discussion}

\setlength{\tabcolsep}{4.5pt}
\begin{table}
\label{table:results}
\centering
\caption{Performances, over 3 runs, of the models trained to recognize entry boundaries, with different experimental setups. The models are sorted in ascending order of the macro F-score (in \%) of the entities EBEGIN and EEND, computed as the harmonic mean of the average precision and recall for both entities over all runs. Finally, the global NER F-score for all named entities is given for experiments 2.1 and 2.2.  Abs. stands for \emph{absolute indents values}, rel. for \emph{absolute indents values}}

\begin{tabular}{l|lllll|ccc|c|}
\cmidrule{2-10}
 & \multicolumn{5}{|c|}{Experimental setup} & \multicolumn{3}{|c|}{EBEGIN \& EEND (\%)} & \\
\rot{\# experiment} & \rot{text} & \rot{ner on text content} & \rot{visual breaks} & \rot{Left spaces} & \rot{right spaces} &                \rot{Precision} & \rot{Recall} & \rot{F-score } & \rot{Overall NER F-score} \\
\cmidrule{2-10}
             2.2 &                yes &            yes &                 yes &                    rel. &                     abs. &                           94.7 &         56.3 &     70.6 &        93.0 \\
             1.1 &                yes &             no &                  no &                      no &                       no &                           97.3 &         95.5 &     96.4 &        - \\
             1.7 &                 no &             no &                 yes &                    rel. &                     abs. &                           98.4 &         97.9 &     98.2 &        - \\
             1.2 &                yes &             no &                 yes &                      no &                       no &                           99.5 &         98.0 &     98.7 &       - \\
             1.4 &                yes &             no &                 yes &                    rel. &                       no &                           99.7 &         98.2 &     98.9 &        - \\
             1.3 &                yes &             no &                 yes &                    abs. &                       no &                           99.7 &         98.3 &     99.0 &        - \\
             1.6 &                yes &             no &                 yes &                    rel. &                     abs. &                           99.4 &         98.6 &     99.0 &        - \\
             1.5 &                yes &             no &                 yes &                    abs. &                     abs. &                           99.6 &         98.5 &     99.0 &        - \\
             2.1 &                yes &            yes &                 yes &                    rel. &                     abs. &                           99.3 &         99.0 &     99.2 &        94.0 \\
\cmidrule{2-10}
\end{tabular}
\end{table}

From the group of experiments 1.X conducted on the entry separation task without NER, several conclusions can be drawn.
Experiment 1.1 proves that text-only entry separation is a solid baseline, reaching an F-score of 96.4\% on the entry separation task from noisy OCR text.
Experiments 1.2 to 1.6 show that progressive integration of visual (structural) information steadily improves the quality of entry separation, reaching an F-score of 99.0\% for the best configuration, composed of line and column breaks and absolute left and right spaces.
This configuration is tied with the variant with relative left spaces, letting us believe that the network captures similar information from absolute and relative variants in these documents.
Experiment 1.7, on the other hand, shows that using structural tokens only forms a solid baseline that reaches an F-score of 98.2\%.
This is due to the short nature of entries in the dataset and the capacity of the network to model sequence state very efficiently, but may not be so significant on datasets with larger units.
This first group proves the feasibility of our proposed approach and the positive interaction visual and textual information can have when projected into the same stream of tokens, processed by a mixed language model.

The group of experiment 2.X conducted on a joint task of entry separation and NER, proves that such unified process can be accomplished by a single network, leveraging the training effort and the computation required to extract named entities.
Experiment 2.1 shows that extra tokens injected into the input stream can also be useful to simplify the output objective of the network:
by trying to label the spacing tokens as entry beginning and end, we avoid the need for the complex join labels used in experiment 2.2 and reach the best absolute performance in terms of entry segmentation with an F-score of 99.2\%, while reaching a good NER performance.
Experiment 2.2, on the other hand, shows that it is possible to annotate only textual tokens, but at the cost of an increase of target labels. Such a system is harder to converge and reaches a lower performance.
We do not report the results for an experiment which would perform NER and entry separation at the same time from text only without joint labeling, as it would require adding another network head and would not make a fair comparison with experiment 2.1 because of the extra weights.

 \section{Conclusion and perspectives}

We have assessed the importance of combining visual and textual information for the particular case of entry separation in 19\textsuperscript{th} century trade directories with raw OCR text.
As expected, by mixing indentation, break, and layout elements with the regular text stream, we outperform the systems that would use only a single of those modalities.
Nevertheless, the originality of this paper lies in the way we have mixed this information. 
While the current trend is to convert text content into visual ``pixel'' content, we choose the opposite way: turning visual clues into text tokens. 
This approach has many advantages.
Compared to visual transformers, it is simpler, lighter to fine-tune using some pre-trained network, and requires less annotated data, which is easier to produce. 
Also, as a NER system is necessary to extract structured information, we maximize computation efficiency by performing two tasks at the same time with this same network, benefiting from richer annotations during training.
Eventually, such system can process an infinite stream of tokens, and reduces the burden of handling content units spanning across lines, columns, or pages.

Code, data, results and models are available at \url{https://github.com/soduco/paper-entryseg-icdar23-code},
\url{https://huggingface.co/HueyNemud/} \\(\texttt{icdar23-entrydetector*} variants). \subsubsection{Acknowledgements} 
This work is supported by the French National Research Agency (ANR), as part of the SODUCO project, under Grant ANR-18-CE38-0013.

\clearpage
\bibliographystyle{splncs04}
\bibliography{nerseg.bib}

\begin{thebibliography}{10}
\providecommand{\url}[1]{\texttt{#1}}
\providecommand{\urlprefix}{URL }
\providecommand{\doi}[1]{https://doi.org/#1}

\bibitem{abadie.das22}
Abadie, N., Carlinet, E., Chazalon, J., Dum{\'e}nieu, B.: A benchmark of named
  entity recognition approaches in historical documents - application to
  19$^{th}$ century french directories. In: Uchida, S., Barney, E., Eglin, V.
  (eds.) Proc. of Document Analysis Systems. DAS 2022. Lecture Notes in
  Computer Science, vol. 13237. Springer, Cham (May 2022).
  \doi{10.1007/978-3-031-06555-2_30}

\bibitem{binmakhashen2019document}
Binmakhashen, G.M., Mahmoud, S.A.: Document layout analysis: a comprehensive
  survey. ACM Computing Surveys (CSUR)  \textbf{52}(6),  1--36 (2019)

\bibitem{chen2017convolutional}
Chen, K., Seuret, M., Hennebert, J., Ingold, R.: Convolutional neural networks
  for page segmentation of historical document images. In: 2017 14th IAPR
  International Conference on Document Analysis and Recognition (ICDAR).
  vol.~1, pp. 965--970. IEEE (2017)

\bibitem{chen2014page}
Chen, K., Wei, H., Hennebert, J., Ingold, R., Liwicki, M.: Page segmentation
  for historical handwritten document images using color and texture features.
  In: 2014 14th International Conference on Frontiers in Handwriting
  Recognition. pp. 488--493. IEEE (2014)

\bibitem{chi2019complicated}
Chi, Z., Huang, H., Xu, H.D., Yu, H., Yin, W., Mao, X.L.: Complicated table
  structure recognition. arXiv preprint arXiv:1908.04729  (2019)

\bibitem{Coquenet2023}
Coquenet, D., Chatelain, C., Paquet, T.: {DAN}: a segmentation-free document
  attention network for handwritten document recognition. {IEEE} Transactions
  on Pattern Analysis and Machine Intelligence pp. 1--17 (2023).
  \doi{10.1109/tpami.2023.3235826}

\bibitem{devlin2018bert}
Devlin, J., Chang, M.W., Lee, K., Toutanova, K.: {BERT}: Pre-training of deep
  bidirectional transformers for language understanding. In: Proc. of
  NAACL-HLT. pp. 4171--4186 (2019)

\bibitem{diem2011text}
Diem, M., Kleber, F., Sablatnig, R.: Text classification and document layout
  analysis of paper fragments. In: 2011 International Conference on Document
  Analysis and Recognition. pp. 854--858. IEEE (2011)

\bibitem{fischer2014combined}
Fischer, A., Baechler, M., Garz, A., Liwicki, M., Ingold, R.: A combined system
  for text line extraction and handwriting recognition in historical documents.
  In: 2014 11th IAPR International Workshop on Document Analysis Systems. pp.
  71--75. IEEE (2014)

\bibitem{garz2010detecting}
Garz, A., Diem, M., Sablatnig, R.: Detecting text areas and decorative elements
  in ancient manuscripts. In: 2010 12th International Conference on Frontiers
  in Handwriting Recognition. pp. 176--181. IEEE (2010)

\bibitem{garz2011layout}
Garz, A., Sablatnig, R., Diem, M.: Layout analysis for historical manuscripts
  using sift features. In: 2011 International Conference on Document Analysis
  and Recognition. pp. 508--512. IEEE (2011)

\bibitem{gruning2019two}
Gr{\"u}ning, T., Leifert, G., Strau{\ss}, T., Michael, J., Labahn, R.: A
  two-stage method for text line detection in historical documents.
  International Journal on Document Analysis and Recognition (IJDAR)
  \textbf{22}(3),  285--302 (2019)

\bibitem{gruning_read-bad_2018}
Grüning, T., Labahn, R., Diem, M., Kleber, F., Fiel, S.: {READ}-{BAD}: {A}
  {New} {Dataset} and {Evaluation} {Scheme} for {Baseline} {Detection} in
  {Archival} {Documents}. In: 2018 13th {IAPR} {International} {Workshop} on
  {Document} {Analysis} {Systems} ({DAS}). pp. 351--356 (Apr 2018).
  \doi{10.1109/DAS.2018.38}

\bibitem{gu2022xylayoutlm}
Gu, Z., Meng, C., Wang, K., Lan, J., Wang, W., Gu, M., Zhang, L.: Xylayoutlm:
  Towards layout-aware multimodal networks for visually-rich document
  understanding. In: Proceedings of the IEEE/CVF Conference on Computer Vision
  and Pattern Recognition. pp. 4583--4592 (2022)

\bibitem{ha1995recursive}
Ha, J., Haralick, R.M., Phillips, I.T.: Recursive xy cut using bounding boxes
  of connected components. In: Proceedings of 3rd International Conference on
  Document Analysis and Recognition. vol.~2, pp. 952--955. IEEE (1995)

\bibitem{hao_language_2022}
Hao, Y., Song, H., Dong, L., Huang, S., Chi, Z., Wang, W., Ma, S., Wei, F.:
  Language models are general-purpose interfaces.
  \doi{10.48550/arXiv.2206.06336}, \url{http://arxiv.org/abs/2206.06336}

\bibitem{harley2015evaluation}
Harley, A.W., Ufkes, A., Derpanis, K.G.: Evaluation of deep convolutional nets
  for document image classification and retrieval. In: 2015 13th International
  Conference on Document Analysis and Recognition (ICDAR). pp. 991--995. IEEE
  (2015)

\bibitem{hong2022bros}
Hong, T., Kim, D., Ji, M., Hwang, W., Nam, D., Park, S.: Bros: A pre-trained
  language model focusing on text and layout for better key information
  extraction from documents. In: Proceedings of the AAAI Conference on
  Artificial Intelligence. vol.~36, pp. 10767--10775 (2022)

\bibitem{huang2022layoutlmv3}
Huang, Y., Lv, T., Cui, L., Lu, Y., Wei, F.: Layoutlmv3: Pre-training for
  document ai with unified text and image masking. In: Proceedings of the 30th
  ACM International Conference on Multimedia. pp. 4083--4091 (2022)

\bibitem{huang2019icdar2019}
Huang, Z., Chen, K., He, J., Bai, X., Karatzas, D., Lu, S., Jawahar, C.:
  Icdar2019 competition on scanned receipt ocr and information extraction. In:
  2019 International Conference on Document Analysis and Recognition (ICDAR).
  pp. 1516--1520. IEEE (2019)

\bibitem{jaume2019funsd}
Jaume, G., Ekenel, H.K., Thiran, J.P.: Funsd: A dataset for form understanding
  in noisy scanned documents. In: 2019 International Conference on Document
  Analysis and Recognition Workshops (ICDARW). vol.~2, pp.~1--6. IEEE (2019)

\bibitem{katti_chargrid_2018}
Kat~ti, A.R., Reisswig, C., Guder, C., Brarda, S., Bickel, S., Höhne, J.,
  Faddoul, J.B.: Chargrid: {Towards} {Understanding} {2D} {Documents} (Sep
  2018), \url{https://arxiv.org/abs/1809.08799v1}

\bibitem{kudo_sentencepiece_2018}
Kudo, T., Richardson, J.: {SentencePiece}: {A} simple and language independent
  subword tokenizer and detokenizer for {Neural} {Text} {Processing}. In:
  Proceedings of the 2018 {Conference} on {Empirical} {Methods} in {Natural}
  {Language} {Processing}: {System} {Demonstrations}. pp. 66--71. Association
  for Computational Linguistics, Brussels, Belgium (Nov 2018).
  \doi{10.18653/v1/D18-2012}, \url{https://aclanthology.org/D18-2012}

\bibitem{li-etal-2021-structurallm}
Li, C., Bi, B., Yan, M., Wang, W., Huang, S., Huang, F., Si, L.:
  {S}tructural{LM}: Structural pre-training for form understanding. In:
  Proceedings of the 59th Annual Meeting of the Association for Computational
  Linguistics and the 11th International Joint Conference on Natural Language
  Processing (Volume 1: Long Papers). pp. 6309--6318. Association for
  Computational Linguistics, Online (Aug 2021).
  \doi{10.18653/v1/2021.acl-long.493},
  \url{https://aclanthology.org/2021.acl-long.493}

\bibitem{li2020}
Li, J., Sun, A., Han, J., Li, C.: A survey on deep learning for named entity
  recognition. IEEE Transactions on Knowledge and Data Engineering
  \textbf{34}(1),  50--70 (2020)

\bibitem{liu-etal-2019-graph}
Liu, X., Gao, F., Zhang, Q., Zhao, H.: Graph convolution for multimodal
  information extraction from visually rich documents. In: Proceedings of the
  2019 Conference of the North {A}merican Chapter of the Association for
  Computational Linguistics: Human Language Technologies, Volume 2 (Industry
  Papers). pp. 32--39. Association for Computational Linguistics, Minneapolis,
  Minnesota (Jun 2019). \doi{10.18653/v1/N19-2005},
  \url{https://aclanthology.org/N19-2005}

\bibitem{mansouri2008}
Mansouri, A., Affendey, L.S., Mamat, A.: Named entity recognition approaches.
  TAL  \textbf{52}(1),  339–344 (2008)

\bibitem{mathew2021docvqa}
Mathew, M., Karatzas, D., Jawahar, C.: Docvqa: A dataset for vqa on document
  images. In: Proceedings of the IEEE/CVF winter conference on applications of
  computer vision. pp. 2200--2209 (2021)

\bibitem{maurel2011}
Maurel, D., Friburger, N., Antoine, J.Y., Eshkol-Taravella, I., Nouvel, D.:
  Casen : a transducer cascade to recognize french named entities. TAL
  \textbf{52}(1),  69–96 (2011)

\bibitem{meunier2005optimized}
Meunier, J.L.: Optimized xy-cut for determining a page reading order. In:
  Eighth International Conference on Document Analysis and Recognition
  (ICDAR'05). pp. 347--351. IEEE (2005)

\bibitem{newseye2021}
Michael, J., Weidemann, M., Laasch, B., Labahn, R.: Icpr 2020 competition on
  text block segmentation on a newseye dataset. In: Pattern Recognition. ICPR
  International Workshops and Challenges: Virtual Event, January 10-15, 2021,
  Proceedings, Part VIII. p. 405–418. Springer-Verlag, Berlin, Heidelberg
  (2021). \doi{10.1007/978-3-030-68793-9_30}

\bibitem{michael_icpr_2021}
Michael, J., Weidemann, M., Laasch, B., Labahn, R.: {ICPR} 2020 {Competition}
  on {Text} {Block} {Segmentation} on a {NewsEye} {Dataset}. In: Del~Bimbo, A.,
  Cucchiara, R., Sclaroff, S., Farinella, G.M., Mei, T., Bertini, M.,
  Escalante, H.J., Vezzani, R. (eds.) Pattern {Recognition}. {ICPR}
  {International} {Workshops} and {Challenges}. pp. 405--418. Lecture {Notes}
  in {Computer} {Science}, Springer International Publishing, Cham (2021).
  \doi{10.1007/978-3-030-68793-9_30}

\bibitem{nadeau2007}
Nadeau, D., Sekine, S.: A survey of named entity recognition and
  classification. Lingvisticae Investigationes  \textbf{30}(1),  3--–26
  (2007)

\bibitem{nagy1984hierarchical}
Nagy, G., Seth, S.C.: Hierarchical representation of optically scanned
  documents. In: " International conference on Pattern Recognition (1984)

\bibitem{seqeval}
Nakayama, H.: {seqeval}: A python framework for sequence labeling evaluation
  (2018), \url{https://github.com/chakki-works/seqeval}

\bibitem{nouvel2011}
Nouvel, D., Antoine, J.Y., Friburger, N., Soulet, A.: Recognizing named
  entities using automatically extracted transduction rules. In: {5th Language
  and Technology Conference}. pp. 136--140. Poznan, Poland (2011)

\bibitem{o1993document}
O'Gorman, L.: The document spectrum for page layout analysis. IEEE Transactions
  on pattern analysis and machine intelligence  \textbf{15}(11),  1162--1173
  (1993)

\bibitem{oliveira2018dhsegment}
Oliveira, S.A., Seguin, B., Kaplan, F.: dhsegment: A generic deep-learning
  approach for document segmentation. In: 2018 16th International Conference on
  Frontiers in Handwriting Recognition (ICFHR). pp. 7--12. IEEE (2018)

\bibitem{park2019cord}
Park, S., Shin, S., Lee, B., Lee, J., Surh, J., Seo, M., Lee, H.: Cord: a
  consolidated receipt dataset for post-ocr parsing. In: Workshop on Document
  Intelligence at NeurIPS 2019 (2019)

\bibitem{ren_faster_2015}
Ren, S., He, K., Girshick, R., Sun, J.: Faster {R}-{CNN}: towards real-time
  object detection with region proposal networks. In: Proceedings of the 28th
  {International} {Conference} on {Neural} {Information} {Processing} {Systems}
  - {Volume} 1. {NIPS}'15, vol.~28, pp. 91--99. MIT Press, Cambridge, MA, USA
  (2015)

\bibitem{sutheebanjard2010modified}
Sutheebanjard, P., Premchaiswadi, W.: A modified recursive xy cut algorithm for
  solving block ordering problems. In: 2010 2nd International Conference on
  Computer Engineering and Technology. vol.~3, pp. V3--307. IEEE (2010)

\bibitem{nerval}
Teklia: Nerval, a python package for {NER} evaluation on noisy text. online
  (2022), \url{https://gitlab.com/teklia/ner/nerval}

\bibitem{vaswani2017attention}
Vaswani, A., Shazeer, N., Parmar, N., Uszkoreit, J., Jones, L., Gomez, A.N.,
  Kaiser, {\L}., Polosukhin, I.: Attention is all you need. In: Adv. Neural
  Inf. Process. Syst. pp. 5998--6008 (2017)

\bibitem{wang-etal-2022-lilt}
Wang, J., Jin, L., Ding, K.: {L}i{LT}: A simple yet effective
  language-independent layout transformer for structured document
  understanding. In: Proceedings of the 60th Annual Meeting of the Association
  for Computational Linguistics (Volume 1: Long Papers). pp. 7747--7757.
  Association for Computational Linguistics, Dublin, Ireland (May 2022).
  \doi{10.18653/v1/2022.acl-long.534},
  \url{https://aclanthology.org/2022.acl-long.534}

\bibitem{xu-etal-2021-layoutlmv2}
Xu, Y., Xu, Y., Lv, T., Cui, L., Wei, F., Wang, G., Lu, Y., Florencio, D.,
  Zhang, C., Che, W., Zhang, M., Zhou, L.: {L}ayout{LM}v2: Multi-modal
  pre-training for visually-rich document understanding. In: Proceedings of the
  59th Annual Meeting of the Association for Computational Linguistics and the
  11th International Joint Conference on Natural Language Processing (Volume 1:
  Long Papers). pp. 2579--2591. Association for Computational Linguistics,
  Online (Aug 2021). \doi{10.18653/v1/2021.acl-long.201},
  \url{https://aclanthology.org/2021.acl-long.201}

\bibitem{xu2020layoutlm}
Xu, Y., Li, M., Cui, L., Huang, S., Wei, F., Zhou, M.: Layoutlm: Pre-training
  of text and layout for document image understanding. In: Proceedings of the
  26th ACM SIGKDD International Conference on Knowledge Discovery \& Data
  Mining. pp. 1192--1200 (2020)

\bibitem{xu2021layoutxlm}
Xu, Y., Lv, T., Cui, L., Wang, G., Lu, Y., Florencio, D., Zhang, C., Wei, F.:
  Layoutxlm: Multimodal pre-training for multilingual visually-rich document
  understanding. arXiv preprint arXiv:2104.08836  (2021)

\bibitem{yang2017learning}
Yang, X., Yumer, E., Asente, P., Kraley, M., Kifer, D., Lee~Giles, C.: Learning
  to extract semantic structure from documents using multimodal fully
  convolutional neural networks. In: Proceedings of the IEEE Conference on
  Computer Vision and Pattern Recognition. pp. 5315--5324 (2017)

\end{thebibliography}
\end{document}